\documentclass[11pt, a4paper, copyright, goog]{google}

\usepackage[authoryear, sort&compress, round]{natbib}
\newcommand{\system}{LEAP}
\newcommand{\systemfull}{LEAP (LLM-in-Lean Environment Agentic Prover)}

\usepackage{hyperref}       
\usepackage{xurl}
\usepackage{url}            
\usepackage{booktabs}       
\usepackage{amsfonts}       
\usepackage{nicefrac}       
\usepackage{microtype}      
\usepackage{xcolor}         
\usepackage{graphicx}
\usepackage[table]{xcolor} 
\usepackage{tabularx}    
\usepackage{multirow}
\usepackage{booktabs}
\usepackage{comment}
\usepackage{enumitem}
\usepackage[most]{tcolorbox}
\tcbuselibrary{listings,breakable}

\tcbset{
  proofartifact/.style={
    breakable,
    colback=gray!3,
    colframe=gray!40,
    fonttitle=\bfseries,
    fontupper=\footnotesize,
  },
  leanartifact/.style={
    listing only,
    breakable,
    colback=gray!3,
    colframe=gray!40,
    fonttitle=\bfseries,
    listing options={
      basicstyle=\ttfamily\footnotesize,
      breaklines=true,
      columns=fullflexible,
      keepspaces=true,
      mathescape=true,
      escapeinside={(*}{*)}
    }
  },
  textartifact/.style={
    proofartifact, 
    text only      
  }
}

\uselogo{} 

\title{\system{}: Supercharging LLMs for Formal Mathematics with Agentic Frameworks}

\correspondingauthor{ponienkung@ucla.edu, violetpeng@google.com}

\reportnumber{} 

\renewcommand{\today}{2026-11-12}

\author[1,4]{Po-Nien Kung}
\author[2]{Linfeng Song}
\author[3]{Dawsen Hwang}
\author[1]{Jinsung Yoon}
\author[1]{Chun-Liang Li}
\author[2]{Simone Severini}
\author[3]{Mirek Olšák}
\author[3]{Edward Lockhart}
\author[3]{Quoc V Le}
\author[1]{Burak Gokturk}
\author[3]{Thang Luong}
\author[1]{Tomas Pfister}
\author[1,4]{Nanyun Peng}

\affil[1]{Google Cloud AI Research}
\affil[2]{Google Cloud}
\affil[3]{Google DeepMind}
\affil[4]{University of California, Los Angeles}

\begin{abstract}
Large language models (LLMs) demonstrate strong informal mathematical reasoning, yet constructing complex, mechanically verifiable proofs remains challenging. We introduce \systemfull, an agentic framework that uses general-purpose LLMs without specialized prover fine-tuning. At its core, \system{} maintains a persistent AND--OR directed acyclic graph (DAG) as structured proof memory, organizing alternative proof sketches and their lemma dependencies. This design enables verified intermediate results to be preserved and reused across proof branches, while iterative compiler feedback and decomposition review guide proof search. Our analysis shows that iterative formalization and decomposition improve proving performance, and that DAG-based memoization yields further gains over tree-structured search. We introduce Lean-IMO-Bench, a benchmark of challenging IMO-style problems formalized in Lean. \system{} solves all 12 Putnam 2025 problems with lower reported aggregate runtime than Numina-Lean-Agent and Axiom, and achieves 83.3\% and 56.7\% on Lean-IMO-Bench's Basic and Advanced sets, respectively. Beyond competition mathematics, \system{} demonstrates research-level capabilities by formalizing a key subproblem of Knuth's Hamiltonian decomposition problem and solving many more research-level problems, including all nine Erd\H{o}s problems solved by AlphaProof Nexus. For more results and ongoing progress on open mathematical problems, see our project website: \url{https://billkunghappy.github.io/LEAP/}.
\end{abstract}

\begin{document}

\maketitle

\section{Introduction}
\label{sec:intro}

Large Language Models (LLMs) have made impressive progress on mathematical reasoning with natural language, also known as ``informal math reasoning'', demonstrating strong capabilities in complex reasoning and problem-solving for both math competitions and research level maths \citep{huang2025winning, luong2025geminiimo, feng2026towards, feng2026aletheia, feng2026semi}. However, as discussed in recent works like Hilbert \citep{hilbert} and Goedel-Prover-V2 \citep{goedel_prover_v2}, solutions in natural language frequently suffer from logical fallacies and hallucinations, and they are hard to automatically verify. This difficulty in verification is not merely a limitation of automated systems; even for human mathematicians, verifying complex proofs is a notoriously time-consuming process requiring scarce expert labor \citep{greiffenhagen2024checking}. A famous example is the proof of the Kepler conjecture \citep{hales2005proof}, which required four years of peer review before the referees could only claim to be ``99\% certain'' of its correctness,\footnote{\url{https://en.wikipedia.org/wiki/Kepler_conjecture}
} eventually necessitating a decade-long formal verification effort \citep{hales2017formal}. This verification bottleneck underscores that assessing correctness is a hard task in natural language, motivating the exploration of formal mathematics, where proofs are written in a machine-checkable language and verified by a rigorous kernel like in Lean~\citep{moura2021lean}, Isabelle \citep{nipkow2002isabelle}, Coq \citep{huet1997coq}, HOL Light \citep{harrison2009hol}, offers automated verification with guaranteed accuracy. Yet, bridging the gap to formal theorem proving remains a significant challenge, and the performance of automated formal provers currently lags substantially behind that of general-purpose LLMs operating in natural language.

To bridge this gap, recent efforts in the research community predominantly fine-tunes specialized prover models (e.g., AlphaProof~\citep{alphaproof}, DeepSeek Prover V2~\citep{deepseek_prover_v2}, Seed Prover~\citep{seed_prover}, Goedel Prover V2~\citep{goedel_prover_v2}) on formal corpora, with the assumption that general LLMs are ineffective for rigorous formal tasks without specialization. Indeed, according to the FormalProofBench~\citep{formalproofbench} and TaoBench~\citep{taobench}, general LLMs often underperform compared to specialized prover models.

Several agentic systems combine general LLMs with specialized proving components. For instance, Hilbert~\citep{hilbert}, AlphaProof Nexus~\citep{sketchprover}, Aristotle~\citep{aristotle}, and Seed Prover V1.5~\citep{seed_prover_v1_5} use general LLMs for informal reasoning but rely on specialized models for Lean proving steps.  
General-purpose LLM agents such as COPRA~\citep{copra}, Numina-Lean-Agent~\citep{numina_lean_agent}, and Archon~\citep{archon} have also demonstrated strong formal proving. Numina-Lean-Agent and Axiom\footnote{\url{https://github.com/AxiomMath/putnam2025} \label{fn:axiom}} both report solving all 12 Putnam 2025 problems. Unlike Numina's flexible coding-agent interface, \system{} maintains an explicit AND--OR DAG of verified decompositions and reusable lemma dependencies.

In this paper, we show that while general LLMs remain limited in one-shot theorem proving, the bottleneck is not language comprehension but generating long, complex, correct proofs in one attempt. General LLMs offer complementary skills to specialized models: strong informal reasoning, instruction following, tool use, and self-refinement. These make them ideal for agentic ATP frameworks, where proof construction is decomposed and iteratively improved.
To this end, we introduce \textbf{\systemfull}, an agentic framework using \emph{only} general LLMs for formal math. Inspired by human mathematical workflows, \system{} decomposes complex goals into proof sketches and supporting lemmas, and maintains them in a persistent AND--OR directed acyclic graph (DAG). Each goal can admit alternative decompositions, while each decomposition requires all of its supporting lemmas to be proved. This structured memory preserves verified intermediate results across branches, enabling lemma reuse, backtracking, and compiler-guided refinement without restarting proof search. \looseness=-1

To evaluate progress beyond saturated benchmarks such as MiniF2F~\citep{zheng2022minif2f} and PutnamBench~\citep{putnambench}, 
we introduce Lean-IMO-Bench, formalizing the challenging informal math benchmark IMO-Bench~\citep{imo-bench} problem statements into Lean. In contrast to existing benchmarks, which either focus on shorter problems or emphasize broad undergraduate coverage, Lean-IMO-Bench targets the complementary regime of elementary statements whose solutions often hinge on highly non-routine insights and unfold through long, multi-step, and structurally intricate proofs, providing a sharper test of formal theorem proving.

Empirically, on the latest 2025 Putnam Competition, a challenging annual undergraduate mathematics competition in North America whose 2025 top score was 110 out of 120 while the median was only 2, \system{} solves all 12 problems in Lean, achieving perfect performance. This matches results reported by Axiom and Numina-Lean-Agent, with lower reported aggregate runtime. 
On Lean-IMO-Bench, \system{} substantially improves general LLMs' solve rate from under 10\% to 70\%, surpassing specialized ATP models (5\%) and Aristotle (48\%), a strong system with specialized ATP components that earned the score for Gold medal at IMO 2025. 
Beyond competition mathematics, \system{} demonstrates its research-level capabilities on a growing collection of mathematical problems. We formalize a key subproblem of Knuth's Hamiltonian decomposition problem and solve all nine Erd\H{o}s problems solved by AlphaProof Nexus~\citep{sketchprover} using Gemini-3.1-Pro, starting from formal problem statements and definitions provided by Formal Conjectures~\citep{firsching2026formalconjecturesopenevolving}, without reference proofs. Further results and ongoing progress on open problems are available on our project website: \url{https://billkunghappy.github.io/LEAP/}.

The contribution of the paper is three-fold: 

\begin{itemize}
    \item \textbf{Structured Agentic Proof Search:} We introduce \system{}, an agentic framework that codifies the human mathematical workflow -- combining high-level blueprint sketching with low-level formal proof generation and iterative compiler feedback. We focus on how explicit proof-state management, verified decomposition, and hierarchical memoization improve the effectiveness of general-purpose LLM agents without task-specific prover fine-tuning.
    \item \textbf{Lean-IMO-Bench Dataset:} To evaluate progress beyond saturated benchmarks such as MiniF2F and PutnamBench, we introduce Lean-IMO-Bench, a new challenging dataset that translates informal problem statements from IMO-Bench into formal Lean statements. Resources are available at \url{https://imobench.github.io}.
    \item \textbf{Competition and Research-Level Results:} \system{} achieves strong results on Lean-IMO-Bench and solves all 12 Putnam 2025 problems with lower reported aggregate runtime than Numina-Lean-Agent and Axiom.\footnote{Lean solutions for Lean-IMO-Bench and Putnam 2025 are available at \url{https://github.com/google-deepmind/superhuman/tree/main/leap}.} Beyond competition mathematics, \system{} demonstrates research-level proving capabilities on a growing collection of mathematical problems, including all nine Erd\H{o}s problems solved by AlphaProof Nexus.
\end{itemize}


\section{\system{}: Blueprint-Driven Automated Theorem Proving}
\label{sec:leap}

\subsection{Formalizing Proofs with Blueprints}

Formalizing mathematical proofs is rarely a one-shot task: it requires a structured plan for progressively translating a high-level argument into Lean. To manage this complexity, recent formalization efforts often use the Lean Blueprint tool~\footnote{\url{https://github.com/PatrickMassot/leanblueprint}}, which let mathematicians write a human-readable proof roadmap linked to Lean code and visualized as a directed acyclic graph (DAG), where each node represents a proof obligation.
This workflow has been instrumental in coordinating large-scale projects such as the formalization roadmap for Fermat's Last Theorem~\footnote{\url{https://leanprover-community.github.io/blog/posts/FLT-announcement/}}, where a multi-year proof effort is organized through an explicit dependency graph.

Inspired by this workflow, we introduce \system{}, an agent for automated theorem proving with hierarchical decomposition and planning. Rather than synthesizing a complete proof in a single pass, \system{} incrementally drafts blueprints, decomposes Lean goals into supporting lemmas, and maintains the evolving proof plan as an AND-OR DAG. 


\begin{figure}[htbp]
    \centering
    \includegraphics[width=0.98\linewidth]{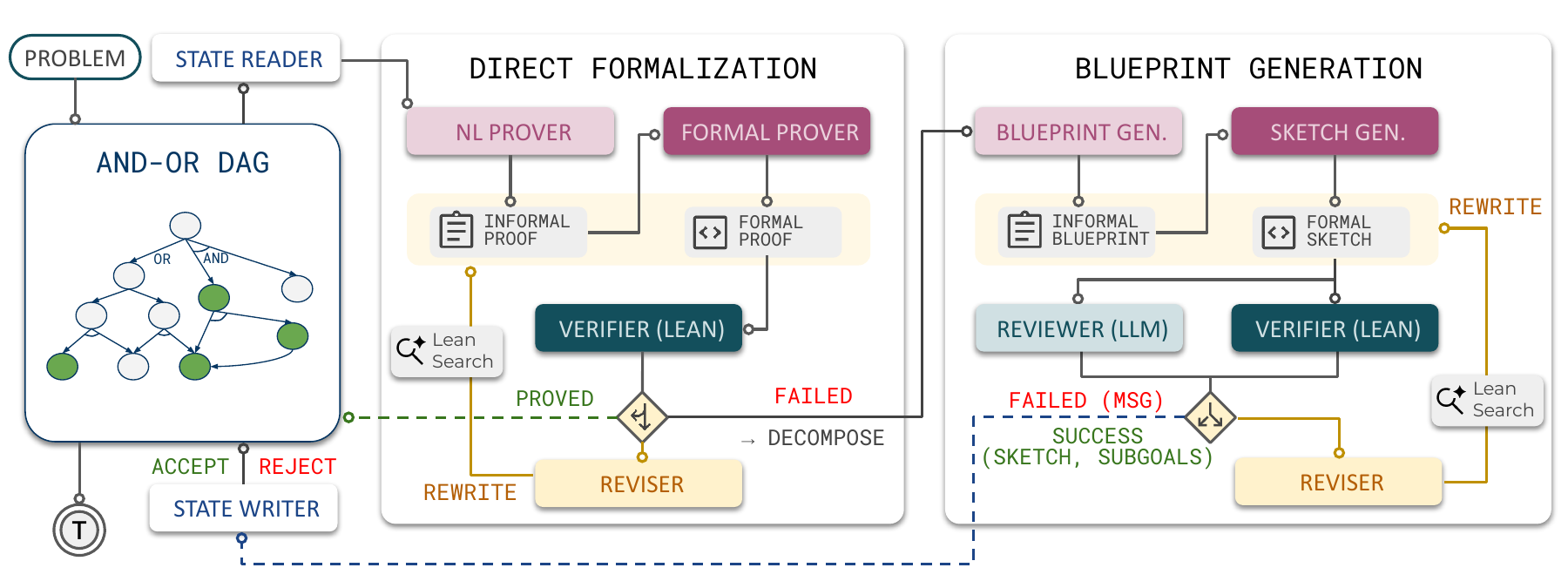}
    \caption{\textbf{\system{} workflow.} \system{} first attempts direct formalization with compiler-feedback revision and LeanSearch\citep{leansearch} retrieval. If this fails, it generates an informal blueprint and formal proof sketch, adding verified subgoals back to the DAG only when dependencies remain acyclic.\looseness=-1}
    \label{fig:workflow}
\end{figure}


\subsection{Overview}



\autoref{fig:workflow} illustrates the workflow of \system{}. Given an input theorem, \system{} registers its Lean statement as the root \textit{goal},
\footnote{A \textit{goal} is any theorem or lemma statement to be proved; decomposition introduces \textit{subgoals}. See \autoref{app:proof_contexts}.} 
represented as an OR node in the AND-OR DAG. To process an open goal, a \textit{state reader} retrieves its statement, dependencies, and related lemmas. \system{} then first attempts a \textbf{direct proof} by generating an informal proof, translating it into Lean code, and checking the candidate with the Lean compiler.

If direct proving fails, \system{} shifts to \textbf{decomposition}. It first drafts an informal blueprint that proposes intermediate lemmas, then translates the blueprint into a Lean proof sketch. The sketch proves the current goal assuming only the proposed lemmas: the main theorem body is \texttt{sorry}-free, while \texttt{sorry} placeholders are permitted only in the newly proposed lemma statements.
If the sketch is accepted by the Lean compiler, it is added as an AND node, and the proposed lemmas are added as child OR nodes. This ensures that once all child subgoals are proved, the parent goal is also proved. The verified sketch is then passed to the \textit{state writer}, which checks that the update preserves acyclicity before committing it to the DAG. The agent then recursively processes the newly created subgoals.

This workflow relies on three tightly coupled design choices: \textbf{DAG-based hierarchical memoization}, which preserves progress and reuses lemmas across branches; \textbf{interleaved informal-formal planning}, which connects natural-language strategies with executable Lean code; and \textbf{verification-guided proof search}, which uses compiler feedback and LLM-based review to accept, revise, decompose, or abandon candidate branches.\looseness=-1

\subsection{Hierarchical Memoization via DAG}

\system{} uses the AND-OR DAG not only to record proof progress, but also to structure hierarchical memoization. OR nodes represent open goals or lemma statements, each of which may be resolved by any valid proof strategy, while AND nodes represent candidate decompositions whose success depends on proving all constituent subgoals. \autoref{fig:dag_example} illustrates this structure.

The DAG provides two central advantages. First, \textbf{monotone refinement}: once a goal is decomposed into supporting subgoals, subsequent search can focus on expanding and resolving these descendants without restructuring the established dependency order. This separates local proof exploration from global proof organization: individual proof attempts may be revised, expanded, or abandoned, while the DAG preserves the stable dependency structure of the overall proof plan. Second, \textbf{lemma memoization}: intermediate lemma statements are stored as shared proof nodes and can be reused whenever the same subproblem arises in different branches. This also supports \textit{anticipatory lemma planning}: during blueprint generation, \system{} may propose auxiliary lemma statements that are not immediately required by the current sketch but could support later proof steps. Such prospective lemmas remain available in the graph memory without being necessary for resolving the current AND node. Together, these properties allow independent proof plans to converge on common dependencies while reducing redundant derivations.

The resulting dependency structure also improves transparency: it exposes which goals remain open, which lemmas have been resolved, and which nodes block downstream progress. This helps \system{} identify where additional lemmas, revised decompositions, or stronger assumptions may be needed, while providing an interpretable blueprint-style workspace for human-AI collaboration.

\subsection{Interleaved Informal--Formal Planning}

As shown in \autoref{fig:workflow}, both the direct proof path and the blueprint decomposition path in \system{} pass through an informal proof sketch. This reflects the complementary strengths of LLMs and Lean: LLMs are effective at informal reasoning, strategy generation, and refinement, while Lean provides strict machine-checkable verification.

In direct proving, \system{} first generates an informal argument for the current goal before translating it into a candidate Lean proof. In decomposition, it drafts an informal blueprint explaining how the goal can be reduced to supporting subgoals, then converts this plan into a Lean sketch that records the proposed dependencies. In both cases, the informal sketch provides a planning space before formalization, making proof construction less brittle than direct code generation alone (see \autoref{app:proof_contexts} for examples of informal proofs and blueprints).

This interleaving also makes proof progress more interpretable: each formal attempt is paired with an informal rationale, allowing users to inspect why a proof step or decomposition was proposed rather than only reading Lean code or compiler feedback.

\subsection{Verification-Guided Proof Search}

As shown in \autoref{fig:workflow}, \system{} uses verification at two levels. First, the Lean compiler formally checks candidate proofs and sketches, ensuring that accepted code is syntactically valid and type-correct. For proof sketches, \system{} only permits \texttt{sorry} placeholders for the proposed subgoals (lemmas). This preserves the AND-OR semantics of the proof DAG: once all referenced subgoals are proved, the parent goal is also proved.
Second, after a blueprint proposes new subgoals, an LLM reviewer assesses the quality of the decomposition: whether the subgoals are relevant to the parent goal, make the problem easier, and offer a plausible route to completing the proof. This planning-level review is crucial for complex goals, where a Lean sketch can be syntactically valid while introducing subgoals that are ill-posed or no simpler than the original statement. Without this filter, the agent may repeatedly expand weak blueprints, spending search budget on branches that do not make real progress. We study this failure mode through an ablation without the LLM reviewer in \autoref{subsec:lm_heuristics}.

The LLM reviewer therefore acts as a search filter: it identifies unpromising decompositions, triggers backtracking, and encourages exploration of alternative strategies. Currently, \system{} uses a simple DFS over the DAG with backtracking. The effectiveness of this reviewer suggests a broader future direction: LLMs may also serve as heuristic evaluators for guiding search in formal proof spaces.

\begin{figure}[htbp]
    \centering
    \includegraphics[width=0.95\linewidth]{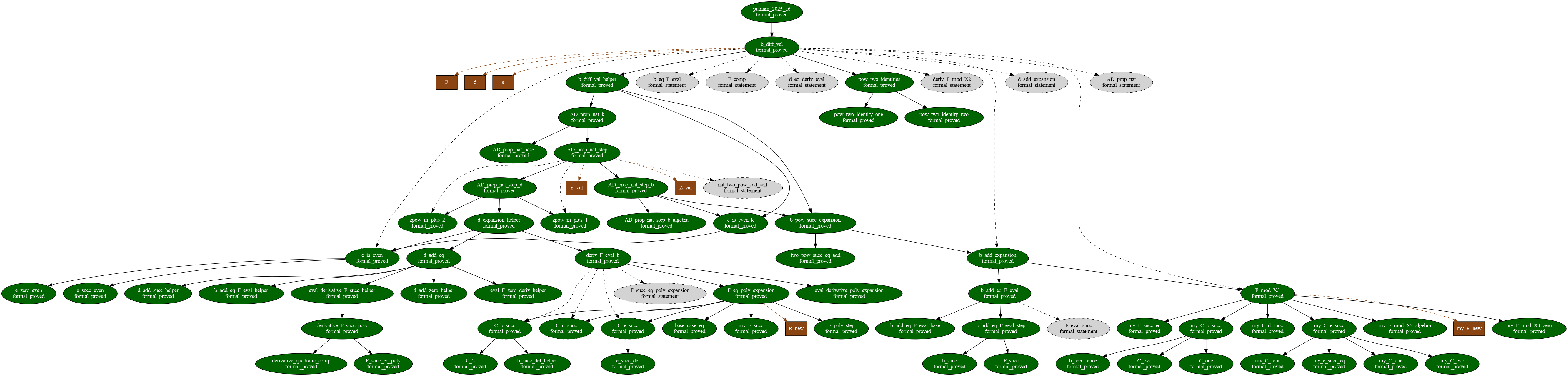}
    \caption{\textbf{DAG example for Putnam 2025 Problem A6.} \system{} decomposes the theorem into a proof sketch and supporting lemmas. Through \textbf{anticipatory lemma planning}, the agent may also propose auxiliary lemma statements that are not immediately required but could be useful later; these are shown with dashed edges and are not needed to prove the main theorem. Green nodes are proven nodes, and brown blocks denote definitions, structures, or variables introduced at a node.} 
    \label{fig:dag_example}
\end{figure}

\section{Lean-IMO-Bench: Formalizing IMO Problems in Lean}
\label{sec:imo-bench}

\begin{table}[ht] \small
\centering
\caption{Baseline performance on \textsc{Lean-IMO-Bench} across three evaluation tasks. Natural Language Proof performance is based on human expert review.}
\label{tab:baseline-results}
\begin{tabular}{@{}llcc@{}}
\toprule
\textbf{Task} & \textbf{Model (Metric)} & \textbf{Basic Set (\%)} & \textbf{Advanced Set (\%)} \\ \midrule
\textbf{Natural Language Proof}   & Gemini 2.5 Pro (Pass@1)      & 55.2 & 17.6 \\ \midrule
\textbf{Formal Theorem Proving}   & Gemini 3.1 Pro (Pass@128)    & 20.0 &  3.3 \\
                                  & Gemini 3.1 Pro (Avg.@128) &  4.6 &  0.2 \\ 
                                 \midrule
\textbf{Formal Proof Translation} & Gemini 3.1 Pro (Pass@128)    & 20.0 &  3.3 \\
                                 & Gemini 3.1 Pro (Avg.@128) &  4.6 &  0.8 \\ 
                                  \bottomrule
\end{tabular}
\end{table}



\subsection{Lean-IMO-Bench}

We introduce \textsc{Lean-IMO-Bench}, a curated collection of 60 problems building upon the foundational work of~\citet{imo-bench}. \citet{imo-bench} introduced \textsc{IMO-ProofBench}, a rigorous suite vetted by an expert panel of mathematicians and IMO medalists. 
The benchmark contains 60 problems split evenly into a \textit{Basic} set and an \textit{Advanced} set of 30 problems each. The \textit{Basic} set spans pre-IMO to IMO-Medium difficulty and includes 8 algebra, 8 combinatorics, 8 number theory, and 6 geometry problems. The \textit{Advanced} set includes novel problems up to IMO-Hard difficulty, with 8 algebra, 8 combinatorics, 6 number theory, and 8 geometry problems. Overall, the benchmark is approximately balanced across algebra, combinatorics, geometry, and number theory.

To ensure the highest level of accuracy in \textsc{Lean-IMO-Bench}, Lean experts manually formalized and verified all 60 problem statements. Because these problems are at the IMO level, the required mathematical background is elementary. Consequently, we expect the corresponding Lean solutions to be concise, deliberately removing the overhead associated with formalizing complex, modern mathematical theories.

The dataset can be used to evaluate models across three distinct tasks: \textbf{Natural Language Proof}, \textbf{Formal Theorem Proving}, and \textbf{Formal Proof Translation}, while we focus on formal theorem proving in this paper. The baseline performance on \textsc{Lean-IMO-Bench} is summarized in \autoref{tab:baseline-results}.
For the Natural Language Proof task,  we cite \citet{imo-bench} as reference: Gemini 2.5 Pro shows strong informal reasoning performance. However, as shown in \autoref{tab:baseline-results}, this does not directly translate to formal theorem proving: Gemini 3.1 Pro performs substantially worse on Formal Theorem Proving, especially on the Advanced set. Providing a correct informal proof in the Formal Proof Translation task also yields little improvement, with Pass@128 unchanged and only a marginal gain in Average@128.

\autoref{tab:baseline-results} demonstrates a stark gap in the models' Lean capabilities. Because the model can already successfully solve these problems in natural language, mathematical reasoning is not the bottleneck, thus reliably generating valid Lean code remains the primary challenge. 


\section{Experimental Results}
\label{sec:experiments}
We evaluate \system{} with Gemini-3.1-pro as the backend large language model and compare it against five baselines: \textbf{Gemini-3.1-pro}, which tests one-shot proof generation by a strong general-purpose model; \textbf{Goedel-Prover-V2-32B}~\citep{goedel_prover_v2}, a state-of-the-art open-source ATP model specialized for Lean; \textbf{Hilbert}~\citep{hilbert}, an agentic Lean formalization framework that combines Goedel-Prover-V2-32B with Gemini-3.1-pro; \textbf{Aristotle}~\citep{aristotle}, a specialized automated theorem-proving system with dedicated ATP components that achieved gold-medal-level performance at the 2025 IMO; and \textbf{Numina-Lean-Agent}~\citep{numina_lean_agent}, a general-purpose coding agent for Lean.

For the headline Putnam 2025 and Lean-IMO-Bench evaluations, \system{} receives only the Lean statements, without reference informal proofs; the Formal Proof Translation baseline in \autoref{tab:baseline-results} does receive reference informal proofs. We evaluate formal proving ability on two datasets: \textbf{Putnam 2025} and our proposed \textbf{Lean-IMO-Bench}. Putnam 2025 contains twelve undergraduate-level problems from the 86th William Lowell Putnam Mathematical Competition,\footnote{Mathematical Association of America, \href{https://maa.org/news/results-of-the-86th-william-lowell-putnam-mathematical-competition/}{\emph{Results of the 86th William Lowell Putnam Mathematical Competition}}.} a highly challenging North American mathematics competition. In the 2025 competition, the top score was 110 out of 120, the average score was approximately 10, and the median score was 2.

\subsection{Results on Putnam 2025}

Table~\ref{tab:putnam_results} presents the Putnam 2025 results. The direct formalization baselines, Gemini-3.1-Pro and Goedel-Prover-V2-32B, fail to solve any problems even at Pass@128, highlighting the difficulty of generating complete, correct proofs without iterative feedback. Agentic methods perform substantially better, demonstrating the value of multi-step proof construction and verification.

\begin{table}[htbp] \small
  \newcommand{\cmark}{\textcolor{green!70!black}{\checkmark}}
  \newcommand{\xmark}{\textcolor{red}{$\times$}}
  \caption{Putnam 2025 results. Checkmarks indicate solved problems. $^\diamond$ denotes Pass@128; $^\dagger$ denotes two rollouts. $^\ddagger$ denotes results reported by the original systems.}
  \label{tab:putnam_results}
  \centering
  \begin{tabularx}{\textwidth}{l *{12}{>{\centering\arraybackslash}X} c}
    \toprule
    Method & a1 & a2 & a3 & a4 & a5 & a6 & b1 & b2 & b3 & b4 & b5 & b6 & Solve Rate (\%) \\
    \midrule
    Gemini-3.1-Pro$^\diamond$ & \xmark & \xmark & \xmark & \xmark & \xmark & \xmark & \xmark & \xmark & \xmark & \xmark & \xmark & \xmark & 0.0 \\
    Goedel-Prover-V2-32B$^\diamond$ & \xmark & \xmark & \xmark & \xmark & \xmark & \xmark & \xmark & \xmark & \xmark & \xmark & \xmark & \xmark & 0.0 \\
    Hilbert$^\dagger$ & \xmark & \cmark & \cmark & \cmark & \xmark & \xmark & \xmark & \xmark & \xmark & \xmark & \xmark & \cmark & 33.3 \\
    Aristotle$^\dagger$ & \cmark & \cmark & \cmark & \cmark & \xmark & \cmark & \cmark & \cmark & \cmark & \cmark & \xmark & \xmark & 75.0 \\
    Numina-Lean-Agent$^\ddagger$ & \cmark & \cmark & \cmark & \cmark & \cmark & \cmark & \cmark & \cmark & \cmark & \cmark & \cmark & \cmark & 100.0 \\
    Axiom$^\ddagger$ & \cmark & \cmark & \cmark & \cmark & \cmark & \cmark & \cmark & \cmark & \cmark & \cmark & \cmark & \cmark & 100.0 \\
    \midrule
    \rowcolor{blue!10}
    \system{}$^\dagger$ & \cmark & \cmark & \cmark & \cmark & \cmark & \cmark & \cmark & \cmark & \cmark & \cmark & \cmark & \cmark & \textbf{100.0} \\
    \bottomrule
  \end{tabularx}
\end{table}

The open-source agentic framework Hilbert solves 4/12 problems, improving considerably over direct generation. Its recursive proof search, however, may repeatedly revisit related subgoals and incur substantial redundant work; we allowed up to seven days per rollout in our evaluation. The proprietary Aristotle system solves 9/12 under two rollouts. Numina-Lean-Agent~\citep{numina_lean_agent} and Axiom also report solving all 12 problems, establishing that perfect Putnam 2025 coverage is achievable by multiple agentic systems.

\system{} likewise solves all 12 problems under two rollouts. Unlike an isolated recursive search, its persistent AND--OR DAG organizes alternative proof sketches, verified subgoals, and reusable intermediate lemmas across the search. This design supports the structured, long-horizon proof search studied in our ablations (\autoref{subsec:dag_vs_tree}), while the Putnam result demonstrates competitive end-to-end proving capability.

\paragraph{Efficiency comparison.} With all three systems achieving 12/12, Table~\ref{tab:putnam_efficiency} compares their reported aggregate runtimes: 822 minutes for \system{}, 2,682 for Axiom, and 3,889 for Numina-Lean-Agent. Token counts are included for reference, but different models and system settings prevent a controlled architecture-level comparison.

\begin{table}[htbp]
  \centering
  \small
  \caption{Per-problem efficiency comparison on Putnam 2025. All three systems solve all 12 problems. Runtime is in minutes; totals sum the reported per-problem runtimes. Numina does not report token counts. Token values are approximate and not directly comparable across backbones and counting conventions. Aggregate token totals are reported separately from the rounded per-problem values.}
  \label{tab:putnam_efficiency}
  \begin{tabularx}{\textwidth}{l *{5}{>{\centering\arraybackslash}X}}
    \toprule
    & \multicolumn{3}{c}{Runtime (min)} & \multicolumn{2}{c}{Reported tokens} \\
    \cmidrule(lr){2-4}\cmidrule(lr){5-6}
    Problem & Numina & Axiom & \system{} & Axiom & \system{} \\
    \midrule
    A1 & 97 & 110 & 13 & 7M & 140K \\
    A2 & 30 & 185 & 59 & 6M & 1.4M \\
    A3 & 44 & 165 & 34 & 8M & 361K \\
    A4 & 169 & 107 & 61 & 8M & 1M \\
    A5 & 2,040 & 518 & 122 & 9.1M & 10M \\
    A6 & 89 & 259 & 90 & 16M & 236K \\
    B1 & 55 & 270 & 113 & 7M & 4.8M \\
    B2 & 142 & 65 & 33 & 2M & 978K \\
    B3 & 30 & 43 & 70 & 2.9M & 1.3M \\
    B4 & 308 & 112 & 61 & 249K & 1.4M \\
    B5 & 88 & 354 & 61 & 18M & 1.5M \\
    B6 & 797 & 494 & 105 & 21M & 2.9M \\
    \midrule
    \textbf{Total} & 3,889 & 2,682 & \textbf{822} & 105.2M & $\sim$28M \\
    \bottomrule
  \end{tabularx}
\end{table}

\subsection{Results on Lean-IMO-Bench}

Table~\ref{tab:imo_results} presents evaluation results on Lean-IMO-Bench. We include this dataset to test model robustness across a broader spectrum of mathematical disciplines and distinct complexity tiers, providing a complementary challenge to the Putnam benchmark.

Direct formalization baselines (Gemini-3.1-Pro and Goedel-Prover-V2-32B) and the open-source Hilbert framework struggle significantly on this dataset, exhibiting severe performance degradation on the Advanced set. While the proprietary Aristotle system resolves a majority of the Basic problems, its effectiveness drops sharply as complexity increases. Notably, across all evaluated methods, performance in the Geometry category remains near zero. This aligns with the well-established difficulty of formalizing olympiad-level geometry in Lean without the aid of supplementary, domain-specific frameworks. We retain this category strictly to evaluate general-purpose reasoning under extreme formalization constraints.

Against these baselines, including Numina-Lean-Agent in its canonical Opus 4.5 configuration (76.7\% Basic / 43.3\% Advanced), \system{} achieves the highest overall solve rates, scoring 83.3\% on the Basic set and 56.7\% on the Advanced set. By effectively leveraging its DAG-based architecture, \system{} demonstrates strong domain generalization, maintaining a 100\% solve rate in both Algebra and Number Theory regardless of the difficulty tier.

\begin{table}[htbp] \small
  \caption{Results on \textsc{Lean-IMO-Bench}. We report the solve rate (\%) across different mathematical categories for the \textbf{Basic} and \textbf{Advanced} sets separately. Evaluation settings: $^\diamond$ indicates pass@128, while $^\dagger$ indicates rollout=2. Numina uses Opus 4.5 with Gemini 3.1 Pro as the informal discussion partner; the Gemini 3.1 Pro variant adapts the harness to a single backbone. Best results in each section are in bold.}
  \label{tab:imo_results}
  \centering
  \small
  \begin{tabularx}{\textwidth}{l *{5}{>{\centering\arraybackslash}X}}
    \toprule
    Method & Algebra & Comb. & Num. Theory & Geometry & \textbf{Overall} \\
    \midrule
    \multicolumn{6}{c}{\textbf{Basic Set (\%)}} \\
    \midrule
    Gemini-3.1-Pro$^\diamond$ & 37.5 & 12.5 & 25.0 & 0 &20.0 \\
    Goedel-V2-32B$^\diamond$    & 37.5 & 0 & 0 & 0 & 10.0 \\
    Hilbert$^\dagger$ & 62.5 & 25	& 50 & 0 & 36.6\\
    Aristotle$^\dagger$ & 75 & \textbf{100} & 100 & \textbf{16.7} & 76.7 \\
    Numina (3.1 Pro)$^\dagger$ & 50 & 50 & 50 & \textbf{16.7} & 43.3 \\
    Numina (Opus 4.5)$^\dagger$ & 87.5 & \textbf{100} & 87.5 & \textbf{16.7} & 76.7 \\
    \rowcolor{blue!10} 
    \system{}$^\dagger$               & \textbf{100} & \textbf{100} & \textbf{100} & \textbf{16.7} & \textbf{83.3} \\
    \midrule
    \multicolumn{6}{c}{\textbf{Advanced Set (\%)}} \\
    \midrule
    Gemini-3.1-Pro$^\diamond$ & 0 & 12.5 & 0 & 0 & 3.3 \\
    Goedel-V2-32B$^\diamond$    & 0 & 0 & 0 & 0 & 0 \\
    Hilbert$^\dagger$ & 12.5 & 0 & 16.6 & 0 & 6.6\\
    Aristotle$^\dagger$ & 37.5 & 12.5 & 33.3 & 0 & 20.0\\
    Numina (3.1 Pro)$^\dagger$ & 50 & 12.5 & 33.3 & 0 & 23.3 \\
    Numina (Opus 4.5)$^\dagger$ & 87.5 & \textbf{25} & 66.6 & 0 & 43.3\\
    \rowcolor{blue!10} 
    \system{}$^\dagger$               & \textbf{100} & \textbf{25} & \textbf{100} & \textbf{12.5} & \textbf{56.7} \\
    \bottomrule
  \end{tabularx}
\end{table}

\section{Discussion}
\label{sec:discussion}





\subsection{Beyond One-Shot Formalization}
\label{subsec:foundation_advantage}

A central motivation of \system{} is that general foundation models can be effective iterative formalizers, even when they are not specialized Lean provers. While specialized provers are trained for formal proof synthesis, general models offer complementary capabilities such as instruction following, long-context reasoning, informal planning, tool use, and feedback-based revision.

To isolate this effect, we evaluate the \textit{Direct Formalization} component labeled in \autoref{fig:workflow} under two settings. In the one-shot setting, each model is evaluated with Pass@128 over independently sampled proof attempts. In the iterative setting, each model receives a single initial attempt and up to 20 compiler-feedback revision steps, yielding a Pass@1 result under a smaller sampling budget. As shown in \autoref{tab:ablation_foundation}, Goedel-Prover-V2-32B does not benefit from this feedback loop, while Gemini-3.1-pro improves substantially from \(20.0\%\) to \(36.6\%\).

This suggests that iterative formalization depends on capabilities beyond local Lean proof synthesis. Interpreting compiler errors, maintaining context, and revising proof attempts over multiple steps can be as important as one-shot formal proving accuracy. These results support using a general foundation model as the reasoning backbone of \system{}, while leaving open the possibility of combining it with specialized provers for local proof generation.

\begin{table}[h]
\centering
\caption{\textbf{One-shot} vs. {\textbf{iterative formalization} performance on Lean-IMO-Bench Basic set.}}
\label{tab:ablation_foundation}
\small
\begin{tabular}{lcc}
\toprule
\textbf{Model} & \textbf{One-shot (Pass@128)} & \textbf{Iterative (Pass@1)} \\
\midrule
Goedel-Prover-V2-32B & 10.0 & 6.6 \\
Gemini-3.1-Pro & 20.0 & 36.6 \\
\bottomrule
\end{tabular}
\end{table}

\subsection{Effect of DAG-Based Memoization}
\label{subsec:dag_vs_tree}

\system{} maintains proof progress as a DAG-based memory rather than an isolated decomposition tree. This allows intermediate lemmas to be stored as shared nodes and reused across branches, while exposing graph context such as existing goals, dependencies, and previously proposed lemmas.

To evaluate this design, we compare \system{} with a tree-structured variant that follows the same workflow but removes global lemma sharing. As shown in \autoref{tab:dag_vs_tree}, the tree variant already substantially outperforms Hilbert~\citep{hilbert}, which achieves 36.6\% and 6.6\% on the Basic and Advanced sets, respectively (\autoref{tab:imo_results}). This indicates that interleaved informal--formal planning and verification-guided search are effective even without DAG-based memoization. The full DAG version further improves performance from 73.3\% to 83.3\% on the Basic set and from 40.0\% to 56.7\% on the Advanced set, showing the benefit of global proof memory.

The improvement is especially pronounced on harder categories, such as Advanced Algebra and Advanced Number Theory, where shared lemmas and graph context are more likely to matter. We attribute this gain to two effects. First, the DAG supports anticipatory lemma planning: auxiliary lemmas proposed at higher-level nodes can later be reused by downstream subgoals (\autoref{fig:dag_example}). Second, repeated subproblems can be shared across branches, avoiding the need to rediscover or reprove the same lemma multiple times. Together, these properties reduce redundant derivations and improve proof search efficiency.


\begin{table}[h]
\centering
\caption{\textbf{DAG memoization ablation.} Solve rate (\%) by category on Lean-IMO-Bench Basic (B)/Advanced (A) sets.}
\label{tab:dag_vs_tree}
\small
\setlength{\tabcolsep}{3pt}
\begin{tabular}{lcccccccccc}
\toprule
\multirow{2}{*}{\textbf{Config.}} 
& \multicolumn{2}{c}{\textbf{Alg.}} 
& \multicolumn{2}{c}{\textbf{Comb.}} 
& \multicolumn{2}{c}{\textbf{NT}} 
& \multicolumn{2}{c}{\textbf{Geo.}} 
& \multicolumn{2}{c}{\textbf{Overall}} \\
\cmidrule(lr){2-3} \cmidrule(lr){4-5} \cmidrule(lr){6-7} \cmidrule(lr){8-9} \cmidrule(lr){10-11}
& \textbf{B} & \textbf{A} 
& \textbf{B} & \textbf{A} 
& \textbf{B} & \textbf{A} 
& \textbf{B} & \textbf{A} 
& \textbf{B} & \textbf{A} \\
\midrule
w/o DAG (Naive Tree)
& 100 & 75
& 75 & 25
& 100 & 66.6
& 0 & 0
& 73.3 & 40.0 \\
Full DAG
& 100 & 100
& 100 & 25
& 100 & 100
& 16.7 & 12.5
& 83.3 & 56.7 \\
\bottomrule
\end{tabular}
\end{table}

\subsection{Toward LLM-Guided Proof Search}
\label{subsec:lm_heuristics}

Compiler verification checks whether a proof sketch is formally well-typed, but not whether its decomposition is useful. A sketch may prove the parent goal from proposed lemmas that are unhelpful, overly difficult, or nearly equivalent to the original goal. In \system{}, the LLM reviewer acts as a local search heuristic: it filters candidate decompositions by judging whether they meaningfully simplify the parent goal before they are committed to the DAG.

We focus this ablation on Putnam 2025 Problem A5 because it is one of the most challenging cases in our evaluation, requiring the longest runtime and two rollouts for \system{} to formalize the proof successfully. Removing the LLM-based decomposition reviewer causes the agent to fail even after eight rollout attempts. This contrast suggests that local LLM review provides a useful search signal: it rejects weak decompositions early, triggers backtracking, and prevents the agent from spending rollouts on branches that do not make substantive progress.
We further inspect decomposition traces from the ablated setting; a representative failure case is shown in \autoref{fig:unproductive_decomposition}.
The decomposition is formally admissible but does not simplify the mathematical state. The agent first unfolds the definitions in the grandparent goal to create an intermediate lemma, then folds them back into a proposed subgoal that is syntactically identical to the original statement. Without semantic review, this duplicate lemma is treated as a new step, causing the agent to repeat the same unproductive decomposition until its search budget is exhausted. This failure highlights the potential of LLM-guided proof search: a reviewer can assess whether a proposed lemma actually advances the proof, prune cyclic or non-simplifying branches, and direct compute toward more promising paths.\looseness=-1



\begin{figure}[t]
\centering
\begin{minipage}{\linewidth}
\begin{tcblisting}{
  leanartifact,
  unbreakable,
  boxsep=0pt,
  top=1pt,
  bottom=1pt,
  left=2pt,
  right=2pt
}
-- Grandparent goal
theorem perms_a_equiv_b_card 
  (m : $\mathbb{N}$) (s : Fin (m + 1) $\rightarrow$ $\mathbb{Z}^\times$) (k : Fin (m + 2)) :
  (*\textcolor{blue}{Nat.card (PermsA m s k) = Nat.card (PermsB m s k)}*) := by
  rw [$\leftarrow$ card_perms_a_eq m s k, $\leftarrow$ card_perms_b_eq m s k]
  exact valid_perms_end_eq_sigma_card m s k

-- Intermediate lemma (unfolding the definitions)
lemma valid_perms_end_eq_sigma_card 
  (m : $\mathbb{N}$) (s : Fin (m + 1) $\rightarrow$ $\mathbb{Z}^\times$) (k : Fin (m + 2)) :
  (*\textcolor{olive}{(ValidPermsEnd (m + 1) s k).card = (Finset.sigma ...).card}*) := by
  rw [card_perms_a_eq m s k, card_perms_b_eq m s k]
  exact card_perms_a_eq_card_perms_b m s k

-- Proposed subgoal (syntactically identical to the grandparent goal)
lemma card_perms_a_eq_card_perms_b 
  (m : $\mathbb{N}$) (s : Fin (m + 1) $\rightarrow$ $\mathbb{Z}^\times$) (k : Fin (m + 2)) :
  (*\textcolor{blue}{Nat.card (PermsA m s k) = Nat.card (PermsB m s k)}*) := by (*\textcolor{red}{sorry}*)
\end{tcblisting}
\end{minipage}
\vspace{-0.5em}
\caption{\textbf{Unproductive decomposition without LLM review.} The proposed subgoal restates the grandparent goal, so the decomposition is formally admissible but does not simplify proof search.}
\vspace{-0.5em}
\label{fig:unproductive_decomposition}
\end{figure}


\subsection{Perspective: General LLMs as Formal Provers: From Zero to Hero}
\label{subsec:zero_to_hero}

As demonstrated by \system{}, the seemingly insurmountable gap between the poor one-shot theorem proving performance of general LLMs and state-of-the-art results can be effectively bridged by a well-designed agentic framework. By shifting the paradigm away from relying solely on small specialized LLMs, we show that the extensive knowledge, instruction following, and self-correction capabilities of foundation models are more than sufficient. When scaffolded correctly, these foundation models can progress from near zero formal math performance to solving highly complex problems. 

While small specialized LLMs lack the overarching agentic capabilities of their foundation counterparts, we acknowledge that they still hold value. A hybrid architecture combining the high-level, structural reasoning of a foundation model with the focused, formal step generation of a fine-tuned specialized model could be a highly effective design pattern. However, exploring this hybrid approach remains outside the scope of this paper, as our primary goal is to highlight the standalone power of general purpose LLMs in an agentic workflow.

\section{Case Studies: Research-Level Formalization}

\paragraph{Hamiltonian Decomposition of Directed Cayley Graphs.}
To evaluate \system{} on a highly complex mathematical task, we targeted a recently solved open problem in combinatorics: the Hamiltonian decomposition of the directed Cayley graph $\Gamma_{m}=Cay(\mathbb{Z}_{m}^{3},\{e_{1},e_{2},e_{3}\})$ for even $m$. Originally posed by Donald Knuth, the problem asks whether the graph's directed arcs can be partitioned into exactly three distinct, spanning Hamiltonian cycles . The informal mathematical proof for the even-case construction is exceptionally intricate, relying on heavy combinatorial analysis and localized defect routing across different layers of the graph.
We focused our formalization efforts on a critical subproblem: rigorously verifying that the 2D planar projection of a single color class's routing dynamics forms an unbroken mathematical cycle of length $m^{2}$. The informal arguments for this specific dynamic span roughly 20 pages of dense piecewise maps, parity-dependent intervals, and complex cross-row transitions. To tackle a formal verification of this magnitude, we deployed \system{}, which successfully decomposed the monolithic informal proof into a granular, highly structured proof graph. By autonomously and systematically resolving the interdependent nodes of this graph, \system{} managed to fully verify the complex cycle-merging dynamics, ultimately synthesizing over 5000 lines of rigorous Lean 4 code to complete the formal proof for this subproblem. Full problem descriptions and informal proofs are available at \url{https://github.com/dpwoodru/knuthCycles/tree/main}.
\paragraph{Formalizing Erdős Problem 457.}
We further tested \system{} on Erd\H{o}s Problem 457, a previously resolved number-theory problem concerning prime divisibility in products of consecutive integers, using the formal statement from Formal Conjectures~\citep{firsching2026formalconjecturesopenevolving}. Given an LLM-generated reconstruction of the known informal proof, \system{} translated and refined the argument into a mechanically verified Lean~4 proof without manual formal-proof engineering.

Formal statements and problem descriptions for the case studies above are provided in \autoref{appendix:problem_statements}.

\paragraph{Erd\H{o}s Problems from AlphaProof Nexus.} We additionally evaluated \system{} with Gemini-3.1-Pro on the nine Erd\H{o}s problems solved by AlphaProof Nexus~\citep{sketchprover}. Starting from the formal statements and necessary definitions in Formal Conjectures~\citep{firsching2026formalconjecturesopenevolving}, without reference proofs or web search, \system{} produced complete, mechanically verified Lean proofs for all nine. This comparison concerns the nine solved instances, rather than the full collection considered by AlphaProof Nexus. For more research-level results and ongoing progress on open problems, see our project website: \url{https://billkunghappy.github.io/LEAP/}. Released Lean solutions are available at \url{https://github.com/google-deepmind/superhuman/tree/main/leap}.

\section{Conclusion and Future Work}
\label{subsec:future_directions}

The success of \system{} suggests that modern general-purpose LLMs possess substantial reasoning capabilities for rigorous domain-specific tasks, provided they are coupled with appropriate structural scaffolding. In formal mathematics, this scaffolding naturally takes the form of proof decomposition and verifier-guided refinement: the model decomposes complex theorems into smaller subgoals, while the Lean compiler checks each formal step. This design provides a structured mechanism for translating informal reasoning into mechanically verified proofs.
A central challenge for future work is how to navigate the resulting proof trees efficiently. As decomposition produces increasingly fine-grained subgoals, the search space can grow rapidly. Future systems should therefore improve branch prioritization, decomposition strategies, and compute allocation across large proof searches. Such advances will be critical for scaling agentic formal proving systems to more complex mathematical problems.\looseness=-1

\section*{Acknowledgements}
We thank Michael P. Brenner, Honghao Lin, David Woodruff, Vahab Mirrokni for providing the informal proof of the even-case of Knuth’s Cycles problem. We would also like to thank Ashley Aragorn Khoo, Paul Lezeau, Calle Sönne, and Moritz Firsching for formalizing the Lean problem statements in Lean-IMO-Bench. We especially thank Moritz Firsching for providing the formal statements of all the Erd\H{o}s problems evaluated in this work through Formal Conjectures~\citep{firsching2026formalconjecturesopenevolving}, and acknowledge the Formal Conjectures project for making these formalized problems available.

\bibliography{main}

\appendix
\section{Limitations}
\label{app:limitations}

\system{} is an initial exploration of using general-purpose LLMs for formal mathematical proving with structured proof memory. Our goal is to study how a strong general model can benefit from an explicit AND-OR DAG that organizes goals, decompositions, and reusable lemmas.

There are several natural directions for extending this work. First, \system{} currently uses a simple depth-first search strategy over the DAG. This keeps the search procedure transparent and allows us to isolate the effect of DAG-based memoization, but more informed search policies could further improve efficiency. For example, future work could explore learned node selection, decomposition ranking, or LLM-guided search heuristics.

Second, the main experiments use Gemini-3.1-Pro. In brief single evaluations on the Lean-IMO-Bench Basic set, \system{} achieves 90.0\% with Opus 4.5 and Gemini 3.1 Pro as a discussion partner, and 86.6\% with Gemini 3.5 Flash, compared with 83.3\% in the main setting. These results do not establish performance on the Advanced set or across repeated runs.

Finally, \system{} focuses on proof planning with general LLMs rather than exhaustive optimization of every ATP component. This leaves room for complementary improvements, such as stronger retrieval, tighter integration with specialized Lean provers, and more advanced proof-state prioritization. We view these directions as promising extensions of the framework.

\section{Related Work}
\label{sec:related_work}

The intersection of machine learning and formal mathematics has seen significant growth, driven by the development of powerful language models and the availability of formalized benchmarks.

\paragraph{Neural Theorem Proving}
Early work in neural theorem proving mainly utilize in-house symbolic engines, such as Metamath \citep{polu2020generative}, MM0 \citep{carneiro2019metamath} or some dedicated formal language for geometry problems \citep{lu2021inter}.
Later work such as mathlib \citep{mathlib}, LeanDojo \citep{leandojo} and MiniF2F \citep{zheng2022minif2f} pioneered the use of LLMs for generative theorem proving in Lean.
They serve as pillar that provide a rich library of known theorems, an interactive environment for step-level search and a descent-level of difficulty evaluation set.
To manage the large search space, HyperTree Proof Search \citep{hypertree} and related Monte Carlo tree search methods \citep{lin2024lean,xin2025bfs} have been explored.
While search-based methods operate at the tactic level, Baldur \citep{baldur} and DeepSeek-prover-v1.5 \citep{xin2024deepseek} explored whole-proof generation, attempting to produce a complete proof in a single step. Another promising direction is guiding formal proof search with informal proofs or sketches. The ``draft, sketch, and prove'' methodology \citep{draftsketchprove} demonstrated that using an informal proof as a blueprint can significantly guide and improve formal theorem proving. Our work, \system{}, builds on this intuition by utilizing general LLMs to generate informal blueprints and iteratively refine formal proofs based on compiler feedback, but without relying on specialized fine-tuned models for the formalization step.

\paragraph{Specialized Prover Models}
Recent breakthroughs have often relied on extensive fine-tuning of large models on formal mathematical corpora. Representative works include AlphaProof \citep{alphaproof}, DeepSeek Prover V2 \citep{deepseek_prover_v2}, Seed Prover \citep{seed_prover}, Kimina Prover \citep{wang2025kimina} and Goedel Prover V2 \citep{goedel_prover_v2}. These models achieve state-of-the-art performance by scaling up training and search on formal systems. However, they require substantial computational resources for training and are highly specialized for specific formal languages. In contrast, \system{} demonstrates that general-purpose LLMs, when placed in a proper agentic environment, can achieve competitive performance without such specialized fine-tuning.

\paragraph{General-Purpose Agentic Provers}
COPRA~\citep{copra} uses a general-purpose LLM with feedback-driven tactic search; Numina-Lean-Agent~\citep{numina_lean_agent} uses a flexible coding agent with Lean tools, and Archon~\citep{archon} formalizes arguments with agentic proof search. \system{} instead studies an explicit, persistent AND--OR proof graph for online, goal-directed decomposition and lemma reuse.

\paragraph{Auto-Formalization}
Auto-formalization, the task of translating natural language mathematics into formal statements and proofs, is a critical bridge between informal and formal reasoning. Early work relied on neural machine translation techniques \citep{wu2022autoformalization}. More recently, LLMs have been used to generate formal statements for training provers at scale, as seen in the auto-formalization pipeline of AlphaProof \citep{alphaproof}. \system{} utilizes the strong auto-formalization capabilities of general LLMs within its agentic harness to bridge the gap between informal blueprints and formal proofs.\looseness=-1

\paragraph{Mathematical Reasoning with LLMs}
Large Language Models have shown impressive progress in solving natural-language mathematical problems, demonstrating strong capabilities in complex reasoning. Recent advancements, such as OpenAI o1 \citep{openai_o1} and DeepSeek R1 \citep{deepseek_r1}, have demonstrated the effectiveness of scaling reinforcement learning for complex mathematical tasks, achieving high scores on competitive benchmarks like AIME. However, direct evaluation of these models on formal theorem proving benchmarks often yields low solve rates, highlighting the gap between informal reasoning and formal verification. \system{} addresses this by leveraging the strong informal reasoning and instruction-following capabilities of general LLMs within an agentic harness, enabling them to interact with the Lean compiler and iteratively self-correct, thus bridging the formalization gap without specialized fine-tuning.

\newpage
\section{Problem Statements}
\label{appendix:problem_statements}
We present the LEAN statements of the open problems that we tested with \system{}.

\paragraph{Hamiltonian Decomposition of Directed Cayley Graphs}
The Hamiltonian decomposition problem for the directed Cayley graph $\Gamma_{m}=Cay(\mathbb{Z}_{m}^{3},\{e_{1},e_{2},e_{3}\})$ asks whether its edges can be partitioned into three distinct directed Hamiltonian cycles. For the even-case construction ($m=2h \ge 10$), the 3D routing dynamics of individual color classes can be analytically projected onto a 2D planar ``round map'' defined on a $\mathbb{Z}_{m} \times \mathbb{Z}_{m}$ grid. The formal statement below encodes the exact operational semantics for the Color 2 subgraph—including its parity-dependent structural defects, coordinate shifts, and piecewise transitions—and asserts that its round map forms a single, unbroken cycle of length $m^2$.

\begin{tcblisting}{
  leanartifact,
  title=\textbf{Lean Statement for the Hamiltonian Decomposition of Directed Cayley Graphs},
  label=box:cayley_graph_statement
}
import Mathlib

set_option autoImplicit false

variable (h : $\mathbb{N}$) (hh : 5 $\leq$ h)

abbrev Fiber2 (h : $\mathbb{N}$) := Fin (2 * h) $\times$ Fin (2 * h)

-- 1. Base Coordinate Definitions
def one2 : Fin (2 * h) := $\langle$1, by omega$\rangle$
def mMinusOne2 : Fin (2 * h) := $\langle$2 * h - 1, by omega$\rangle$
def mMinusTwo2 : Fin (2 * h) := $\langle$2 * h - 2, by omega$\rangle$

def succ2c (x : Fin (2 * h)) : Fin (2 * h) := x + one2 h hh
def pred2c (x : Fin (2 * h)) : Fin (2 * h) := x - one2 h hh

-- 2. Exceptional Set Logic (Defects)
def y2SwitchRow (x : Fin (2 * h)) : Prop :=
  x.val = h + 1 $\vee$ x.val = h + 2 $\vee$ x.val = h + 3

instance (x : Fin (2 * h)) : Decidable (y2SwitchRow h x) := by
  unfold y2SwitchRow
  infer_instance

def y2star (x : Fin (2 * h)) : Fin (2 * h) :=
  if y2SwitchRow h x then
    if h % 2 = 0 then mMinusTwo2 h hh else mMinusOne2 h hh
  else
    $\langle$2 * h - 1 - x.val, by omega$\rangle$

def A2 (x : Fin (2 * h)) : Fin (2 * h) :=
  succ2c h hh (y2star h hh x)

def activeB2 (x y : Fin (2 * h)) : Prop :=
  if h % 2 = 0 then
    (x.val = h + 1 $\wedge$ y.val $\leq$ h - 1) $\vee$
      (x.val = h + 4 $\wedge$ h - 3 $\leq$ y.val $\wedge$ y.val $\leq$ 2 * h - 2)
  else
    (x.val = h + 1 $\wedge$ 1 $\leq$ y.val $\wedge$ y.val $\leq$ h - 1) $\vee$
      (x.val = h + 4 $\wedge$ h - 3 $\leq$ y.val)

instance (x y : Fin (2 * h)) : Decidable (activeB2 h x y) := by
  unfold activeB2
  infer_instance

-- 3. The Round Map
def r2Map (p : Fiber2 h) : Fiber2 h :=
  let x := p.1
  let u := pred2c h hh p.2
  if u = A2 h hh x then
    (succ2c h hh x,
      if x.val = h + 1 $\vee$ x.val = h + 2 then u else pred2c h hh u)
  else if activeB2 h x u then
    (x, pred2c h hh u)
  else
    (x, u)

-- 4. The Self-Contained Goal
/-- The unrolled Hamiltonicity goal for the Color 2 round map. -/
theorem color2_singleCycle_unrolled (h6 : 6 $\leq$ h) :
    ($\forall$ p : Fin (2 * h) $\times$ Fin (2 * h), (r2Map h hh)^[(2 * h) * (2 * h)] p = p) $\wedge$
    ($\forall$ (p : Fin (2 * h) $\times$ Fin (2 * h)) (k : $\mathbb{N}$), 0 < k $\rightarrow$ k < (2 * h) * (2 * h) $\rightarrow$ (r2Map h hh)^[k] p $\neq$ p) := by
  (*\textcolor{red}{sorry}*)
\end{tcblisting}

\paragraph{Erdős 457}
Erdős Problem 457 is a number theory challenge concerning the prime divisors of consecutive integers. Specifically, it conjectures the existence of a real number $\varepsilon > 0$ such that for infinitely many integers $n$, every prime number $p \le (2 + \varepsilon)\log n$ divides the product of the $\lfloor\log n\rfloor$ consecutive integers starting from $n+1$. The Lean formalization below, from Formal Conjectures~\citep{firsching2026formalconjecturesopenevolving}, captures this exact asymptotic prime divisibility condition.

\begin{tcblisting}{
  leanartifact,
  title=\textbf{Lean Statement for Erdős Problem 457},
  label=box:erdos_457_statement
}

import Mathlib

theorem erdos_457 : $\exists$ $\varepsilon$ > (0 : $\mathbb{R}$),
    { (n : $\mathbb{N}$) | $\forall$ (p : $\mathbb{N}$), p $\leq$ (2 + $\varepsilon$) * Real.log n $\rightarrow$ p.Prime $\rightarrow$
      p $\mid$ $\prod$ i $\in$ Finset.Icc 1 $\lfloor$Real.log n$\rfloor_+$, (n + i) }.Infinite := by
  (*\textcolor{red}{sorry}*)

\end{tcblisting}

\newpage
\section{Proof Contexts and Artifacts}
\label{app:proof_contexts}

This section describes the formal and informal artifacts used by \system{} during proof planning. Formal artifacts correspond to Lean-level objects that are checked by the compiler or represented in the proof DAG, while informal artifacts correspond to natural-language planning objects used to guide direct proving and decomposition.

\paragraph{Formal context.}
A \textit{proof goal} is a Lean theorem or lemma statement that remains to be proved. The original input theorem is the root proof goal, while lemma statements introduced by decomposition become subgoals in the proof DAG. A \textit{formal proof} is a complete Lean proof of the current proof goal that does not rely on newly proposed unproven lemmas; once accepted by the Lean compiler, the corresponding goal is marked as resolved. A \textit{proof sketch} is a Lean artifact that proves the current goal assuming a set of proposed lemmas. In \system{}, a proof sketch may contain \texttt{sorry} placeholders only for these explicitly proposed lemmas. Thus, a verified proof sketch defines a valid decomposition: once all referenced proposed lemmas are later proved, the current goal is also proved. We present examples of these context using Lean-IMO-Bench Problem 001 and 009 in the Basic Set. \textbf{All artifacts, except for the Proof Goal of the root problem, are created by \system{} automatically.)}

\begin{tcblisting}{
  leanartifact,
  title=\textbf{Example Proof Goal (Lean-IMO-Bench, Basic 001)},
  label=box:proof_goal_example
}
theorem PBBasic001 : {f : $\mathbb{Z}$ $\rightarrow$ $\mathbb{Z}$ | $\forall$ x y, f (2 * x) + 2 * f y = f (f (x + y))}
      = {0} $\cup$ {(fun x $\mapsto$ 2 * x + c)| (c : $\mathbb{Z}$)} :=
by (*\textcolor{red}{sorry}*)
\end{tcblisting}

\begin{tcblisting}{
  leanartifact,
  title=\textbf{Example Formal Proof (Lean-IMO-Bench, Basic 001)},
  label=box:formal_proof_example
}
import Mathlib

theorem PBBasic001 : {f : $\mathbb{Z}$ $\rightarrow$ $\mathbb{Z}$ | $\forall$ x y, f (2 * x) + 2 * f y = f (f (x + y))}
      = {0} $\cup$ {(fun x $\mapsto$ 2 * x + c)| (c : $\mathbb{Z}$)} :=
by
  ext f
  simp only [Set.mem_setOf_eq, Set.mem_union, Set.mem_singleton_iff, Set.mem_range]
  constructor
  $\cdot$ intro h
    have h1 : $\forall$ y, f (f y) = 2 * f y + f 0 := by
      -- [Proof details omitted for brevity]
    have h2 : $\forall$ x, f (2 * x) = 2 * f x - f 0 := by
      -- [Proof details omitted]
    have h3 : $\forall$ x y, f (x + y) = f x + f y - f 0 := by
      -- [Proof details omitted]
      
    have hc : $\exists$ c, c = f 0 := $\langle$f 0, rfl$\rangle$
    rcases hc with $\langle$c, hc_eq$\rangle$
    have hk : $\exists$ k, k = f 1 - c := $\langle$f 1 - c, rfl$\rangle$
    rcases hk with $\langle$k, hk_eq$\rangle$
    
    -- ... [Induction steps for h4 omitted] ...
      
    have h5 : $\forall$ x : $\mathbb{Z}$, f x = k * x + c := by
      -- ... [Negative cases to prove linear form omitted] ...
        
    have eq_all : $\forall$ x y : $\mathbb{Z}$, k * (2 * x) + c + 2 * (k * y + c) = k * (k * (x + y) + c) + c := by
      intro x y
      have h_orig := h x y
      simp only [h5] at h_orig
      exact h_orig
      
    have hk_eq : k * (k - 2) = 0 := by
      -- [Algebraic simplification using eq_all 1 0 and eq_all 0 0 omitted]
        
    have hk2 : k = 0 $\vee$ k = 2 := by
      cases mul_eq_zero.mp hk_eq with
      | inl h1 => left; exact h1
      | inr h2 => right; omega
      
    rcases hk2 with hk0 | hk2
    $\cdot$ left
      have eq00 := eq_all 0 0
      rw [hk0] at eq00
      have hc0 : c = 0 := by linarith [eq00]
      ext x
      simp only [Pi.zero_apply]
      have hfx := h5 x
      rw [hk0, hc0] at hfx
      omega
    $\cdot$ right
      use c
      ext x
      have hfx := h5 x
      rw [hk2] at hfx
      omega

  -- Prove that the derived candidates are indeed solutions mappings
  $\cdot$ rintro (rfl | $\langle$c, rfl$\rangle$)
    $\cdot$ intro x y
      simp only [Pi.zero_apply, mul_zero, add_zero]
    $\cdot$ intro x y
      dsimp only
      ring
\end{tcblisting}

\begin{tcblisting}{
  leanartifact,
  title=\textbf{Example Proof Sketch (Lean-IMO-Bench, Basic 006)},
  label=box:proof_sketch_example
}
import Mathlib

open Polynomial

-- Shared Definitions from the file environment
def K (c : $\mathbb{N}$ $\rightarrow$ $\mathbb{Z}$) : $\mathbb{Z}$ := (c 1)^2 - (2 : $\mathbb{Z}$) * (c 0) * (c 2)

def k_target (c : $\mathbb{N}$ $\rightarrow$ $\mathbb{Z}$) : $\mathbb{N}$ := max 2 (Int.toNat (K c + (1 : $\mathbb{Z}$)))

def esymm_one_target (s : Multiset $\mathbb{R}$) : $\mathbb{N}$ $\rightarrow$ $\mathbb{R}$
  | 0 => 0
  | k + 1 => s.esymm k

def esymm_two_target (s : Multiset $\mathbb{R}$) : $\mathbb{N}$ $\rightarrow$ $\mathbb{R}$
  | 0 => 0
  | 1 => 0
  | k + 2 => s.esymm k

-- Supporting Lemmas (with sorry)
lemma root_count_bound_implies_eq (c : $\mathbb{N}$ $\rightarrow$ $\mathbb{Z}$) (hc : c 0 $\neq$ 0) (k : $\mathbb{N}$)
  (h_not_less : $\neg$ ((($\sum$ i $\in$ Finset.Icc 0 k, monomial i (c i)).rootSet $\mathbb{R}$).ncard < k)) :
  (($\sum$ i $\in$ Finset.Icc 0 k, monomial i (c i)).rootSet $\mathbb{R}$).ncard = k $\wedge$ 
  ($\sum$ i $\in$ Finset.Icc 0 k, monomial i (c i)).natDegree = k := by
  (*\textcolor{red}{sorry}*)

lemma k_le_K_of_eq (c : $\mathbb{N}$ $\rightarrow$ $\mathbb{Z}$) (hc : c 0 $\neq$ 0) (k : $\mathbb{N}$) (hk : (2 : $\mathbb{N}$) $\leq$ k)
  (h_eq : (($\sum$ i $\in$ Finset.Icc 0 k, monomial i (c i)).rootSet $\mathbb{R}$).ncard = k)
  (h_deg : ($\sum$ i $\in$ Finset.Icc 0 k, monomial i (c i)).natDegree = k) :
  (k : $\mathbb{Z}$) $\leq$ K c := by
  (*\textcolor{red}{sorry}*)

lemma k_target_ge_two (c : $\mathbb{N}$ $\rightarrow$ $\mathbb{Z}$) : (2 : $\mathbb{N}$) $\leq$ k_target c := by
  (*\textcolor{red}{sorry}*)

lemma k_target_gt_K (c : $\mathbb{N}$ $\rightarrow$ $\mathbb{Z}$) : K c < (k_target c : $\mathbb{Z}$) := by
  (*\textcolor{red}{sorry}*)

-- Main Theorem
theorem PBBasic006 (c : $\mathbb{N}$ $\rightarrow$ $\mathbb{Z}$) (hc : c 0 $\neq$ 0) : 
  $\exists$ k, (($\sum$ i $\in$ Finset.Icc 0 k, monomial i (c i)).rootSet $\mathbb{R}$).ncard < k := by
  use k_target c
  by_contra h
  have h1 := root_count_bound_implies_eq c hc (k_target c) h
  have h2 := k_le_K_of_eq c hc (k_target c) (k_target_ge_two c) h1.1 h1.2
  have h3 := k_target_gt_K c
  linarith
\end{tcblisting}

\paragraph{Informal context.}
An \textit{informal proof} is a natural-language plan for proving the current goal directly, without introducing a decomposition into new subgoals. It guides the generation of a complete formal proof. An \textit{informal blueprint} is a higher-level natural-language plan that explains how the current goal can be reduced to useful supporting lemmas. Unlike an informal proof, an informal blueprint may introduce proposed lemmas that are not immediately proved and may include auxiliary lemmas for later proof steps. These proposed lemmas are then translated into formal subgoals and organized in the proof DAG. Examples are shown below:

\begin{tcolorbox}[
  textartifact,
  title=\textbf{Example Informal Proof (Lean-IMO-Bench, Basic 001)},
  label=box:informal_proof_example
]
Let $S = \{f : \mathbb{Z} \to \mathbb{Z} \mid \forall x, y \in \mathbb{Z}, f(2x) + 2f(y) = f(f(x+y))\}$ and $T = \{0\} \cup \{x \mapsto 2x + c \mid c \in \mathbb{Z}\}$. We want to show that $S = T$.\\
\\
\textbf{Part 1: Proof that $T \subseteq S$}\\
\\
Suppose $f \in T$. Then either $f$ is the constant zero function ($f = 0$) or $f(x) = 2x + c$ for some $c \in \mathbb{Z}$.\\
\begin{itemize}
  \item Case 1: If $f = 0$, then for any $x, y \in \mathbb{Z}$, the left-hand side is $f(2x) + 2f(y) = 0 + 2(0) = 0$. The right-hand side is $f(f(x+y)) = f(0) = 0$. The two sides match, so $f \in S$.
  \item Case 2: If $f(x) = 2x + c$ for some $c \in \mathbb{Z}$, then for any $x, y \in \mathbb{Z}$, the left-hand side is:
  $f(2x) + 2f(y) = (2(2x) + c) + 2(2y + c) = 4x + 4y + 3c$.
  The right-hand side is:
  $f(f(x+y)) = f(2(x+y) + c) = 2(2x + 2y + c) + c = 4x + 4y + 3c$.
  The two sides are equal, so $f \in S$.
\end{itemize}
This proves $T \subseteq S$.\\
\\
\textbf{Part 2: Proof that $S \subseteq T$}\\
\\
Let $f \in S$. Then for all $x, y \in \mathbb{Z}$, we have:\\
(1) $f(2x) + 2f(y) = f(f(x+y))$.\\
\\
Substitute $x = 0$ into Eq. (1):\\
$f(0) + 2f(y) = f(f(y))$.\\
Let $c = f(0)$. Then for all $y \in \mathbb{Z}$, we have:\\
(2) $f(f(y)) = 2f(y) + c$.\\
\\
Using Eq. (2) on the right-hand side of Eq. (1) (since it holds for any input, including $x+y$), we obtain:\\
(3) $f(2x) + 2f(y) = 2f(x+y) + c$.\\
\\
Substitute $y = 0$ into Eq. (3):\\
$f(2x) + 2f(0) = 2f(x) + c$.\\
Since $f(0) = c$, this becomes $f(2x) + 2c = 2f(x) + c$, which yields:\\
(4) $f(2x) = 2f(x) - c$.\\
\\
Now, substitute Eq. (4) back into the left-hand side of Eq. (3):\\
$(2f(x) - c) + 2f(y) = 2f(x+y) + c$.\\
Rearranging the terms gives:\\
$2f(x+y) = 2f(x) + 2f(y) - 2c$.\\
Because the codomain is $\mathbb{Z}$, we can divide both sides by $2$ to obtain:\\
(5) $f(x+y) = f(x) + f(y) - c$.\\
\\
Define a new function $g: \mathbb{Z} \to \mathbb{Z}$ by $g(x) = f(x) - c$. Then we can rewrite Eq. (5) as:\\
$g(x+y) + c = (g(x) + c) + (g(y) + c) - c$.\\
Simplifying this yields Cauchy's functional equation:\\
$g(x+y) = g(x) + g(y)$ for all $x, y \in \mathbb{Z}$.\\
\\
Since $g$ satisfies $g(x+y) = g(x) + g(y)$ over integers, it must be a linear function. Let $k = g(1) \in \mathbb{Z}$. By standard induction, we have $g(x) = kx$ for all integers $x \ge 0$ (with $g(0) = 0$). For negative integers, $0 = g(0) = g(x + (-x)) = g(x) + g(-x)$, meaning $g(x) = -g(-x) = -(-kx) = kx$. Thus $g(x) = kx$ for all $x \in \mathbb{Z}$. \\
As a result, we have $f(x) = kx + c$ for all $x \in \mathbb{Z}$.\\
\\
To determine the permissible values of $k$ and $c$, substitute $f(x) = kx + c$ back into the original Eq. (1):\\
Left-hand side: $f(2x) + 2f(y) = k(2x) + c + 2(ky + c) = 2kx + 2ky + 3c$.\\
Right-hand side: $f(f(x+y)) = k(f(x+y)) + c = k(k(x+y) + c) + c = k^2x + k^2y + (k+1)c$.\\
\\
For the equality $2kx + 2ky + 3c = k^2x + k^2y + (k+1)c$ to hold for all $x, y \in \mathbb{Z}$, the corresponding coefficients must match. \\
Setting $x=0$ and $y=0$ yields:\\
(6) $3c = (k+1)c$.\\
Setting $x=1$ and $y=0$ yields:\\
$2k + 3c = k^2 + (k+1)c$.\\
Subtracting Eq. (6) from this gives $2k = k^2$, which simplifies to $k(k-2) = 0$. Since $k \in \mathbb{Z}$, the only solutions are $k = 0$ or $k = 2$.

\begin{itemize}
  \item Case A: $k = 0$.
  Substitute $k = 0$ into Eq. (6):
  $3c = c \implies 2c = 0 \implies c = 0$.
  Thus, $f(x) = 0x + 0 = 0$, meaning $f$ is the zero function. Hence $f \in \{0\} \subseteq T$.

  \item Case B: $k = 2$.
  Substitute $k = 2$ into Eq. (6):
  $3c = 3c$, which is true for any $c \in \mathbb{Z}$.
  Thus, $f(x) = 2x + c$ for some $c \in \mathbb{Z}$. Hence $f \in \{x \mapsto 2x + c \mid c \in \mathbb{Z}\} \subseteq T$.
\end{itemize}

In all cases, any function $f \in S$ is also in $T$, proving $S \subseteq T$.\\
Since both subset inclusions have been established, $S = T$.
\end{tcolorbox}

\begin{tcolorbox}[
  textartifact,
  title=\textbf{Example Informal Blueprint (Lean-IMO-Bench, Basic 006)},
  label=box:informal_blueprint_example
]
\textbf{High-Level Mathematical Idea}

The goal is to prove that $k \le K(c)$, where $K(c) = c_1^2 - 2c_0 c_2$, given that the polynomial $P(x) = \sum_{i=0}^k c_i x^i$ with integer coefficients has degree $k$ and exactly $k$ real roots.

The proof elegantly uses multiset symmetric polynomials and the AM-GM inequality, avoiding rational functions or polynomial derivatives:
\begin{enumerate}
  \item \textbf{Polynomial Roots and Splitting}: Since $P$ has natural degree $k$ and $k$ distinct real roots, it splits completely over $\mathbb{R}$, and the multiset of its roots $S$ has cardinality $k$.
  
  \item \textbf{Vieta's Formulas}: Vieta's formulas express the coefficients $c_0, c_1, c_2$ in terms of the elementary symmetric polynomials of $S$:\\
  $c_0 = c_k (-1)^k E_k$\\
  $c_1 = c_k (-1)^{k-1} E_{k-1}$\\
  $c_2 = c_k (-1)^{k-2} E_{k-2}$\\
  where $E_i = \text{esymm}_i(S)$.
  
  \item \textbf{Multiset Identity}: We construct a new multiset $Y$ by mapping each root $x \in S$ to the product of all other roots, $(S \setminus \{x\})$.\texttt{prod}. Purely multiset inductive identities show:\\
  $\sum Y = E_{k-1}$\\
  $\text{esymm}_2(Y) = E_k E_{k-2}$\\
  $\prod Y = (E_k)^{k-1}$
  
  \item \textbf{Sum of Squares}: For the multiset $Z = c_k Y$, we evaluate the sum of its squares $W = \{ z^2 \mid z \in Z \}$.\\
  By the relation $(\sum Z)^2 = \sum (Z^2) + 2 \text{esymm}_2(Z)$, we can algebraically compute $\sum W = c_1^2 - 2c_0 c_2 = K(c)$.
  
  \item \textbf{Integer Product Bound}: The product of $Z$ evaluates to $c_k ((-1)^k c_0)^{k-1}$. Since $P$ has degree $k$ and $c_0 \neq 0$, both $c_0$ and $c_k$ are non-zero integers. Thus, the product of $Z$ is a non-zero integer, implying $\prod W = (\prod Z)^2 \ge 1$.
  
  \item \textbf{AM-GM Inequality}: Applying the AM-GM inequality to the multiset $W$ (which consists of $k$ non-negative real numbers whose product is $\ge 1$), we obtain $\sum W \ge k \implies K(c) \ge k$.
\end{enumerate}

\vspace{0.5em}
\textbf{Required Global Definitions, Variables, or Structures}

No new definitions, axioms, or structures are needed. We use purely standard Mathlib components (like \texttt{Multiset}, \texttt{Polynomial}, and \texttt{esymm}).

\vspace{0.5em}
\textbf{Smaller Lemmas (Subproblems)}

\begin{quote}
\ttfamily\small
lemma coeff\_of\_sum\_Icc (c : $\mathbb{N}$ $\rightarrow$ $\mathbb{Z}$) (k : $\mathbb{N}$) (hk : (2 : $\mathbb{N}$) $\le$ k) (i : $\mathbb{N}$) (hi : i $\le$ k) :\\
\hspace*{2em}($\sum$ j $\in$ Finset.Icc 0 k, Polynomial.monomial j (c j)).coeff i = c i
\end{quote}
\textbf{Purpose}: Simplifies polynomial coefficient extraction from the given sum format.

\begin{quote}
\ttfamily\small
lemma card\_roots\_eq\_of\_ncard\_rootSet \{k : $\mathbb{N}$\} \{P : Polynomial $\mathbb{R}$\}\\
\hspace*{2em}(h\_deg : P.natDegree = k)\\
\hspace*{2em}(h\_ncard : (P.rootSet $\mathbb{R}$).ncard = k) :\\
\hspace*{2em}P.roots.card = k
\end{quote}
\textbf{Purpose}: Shows that a polynomial with $k$ distinct real roots and degree $k$ has exactly $k$ roots counting multiplicities.

\begin{quote}
\ttfamily\small
lemma multiset\_map\_erase\_prod\_sum \{R : Type*\} [CommRing R] (s : Multiset R) :\\
\hspace*{2em}(s.map (fun x => (s.erase x).prod)).sum = s.esymm (s.card - 1)
\end{quote}
\textbf{Purpose}: Multiset identity relating the sum of element-wise excluded products to $E_{k-1}$.

\begin{quote}
\ttfamily\small
lemma multiset\_map\_erase\_prod\_esymm\_two \{R : Type*\} [CommRing R] (s : Multiset R) :\\
\hspace*{2em}(s.map (fun x => (s.erase x).prod)).esymm 2 = s.prod * s.esymm (s.card - 2)
\end{quote}
\textbf{Purpose}: Multiset identity relating the 2nd elementary symmetric polynomial of excluded products to $E_k E_{k-2}$.

\begin{quote}
\ttfamily\small
lemma multiset\_sum\_sq\_eq \{R : Type*\} [CommRing R] (s : Multiset R) :\\
\hspace*{2em}(s.map (fun x => x\textasciicircum 2)).sum = (s.sum)\textasciicircum 2 - (2 : R) * s.esymm 2
\end{quote}
\textbf{Purpose}: Expresses the sum of squares of a multiset in terms of its sum and its 2nd elementary symmetric polynomial.

\begin{quote}
\ttfamily\small
lemma multiset\_map\_erase\_prod\_prod \{R : Type*\} [CommRing R] (s : Multiset R) :\\
\hspace*{2em}(s.map (fun x => (s.erase x).prod)).prod = s.prod \textasciicircum{} (s.card - 1)
\end{quote}
\textbf{Purpose}: Computes the full product of the excluded products multiset.

\begin{quote}
\ttfamily\small
lemma multiset\_sum\_ge\_card\_of\_prod\_ge\_one (W : Multiset $\mathbb{R}$) (hw : $\forall$ x $\in$ W, 0 $\le$ x) (hp : (1 : $\mathbb{R}$) $\le$ W.prod) :\\
\hspace*{2em}(W.card : $\mathbb{R}$) $\le$ W.sum
\end{quote}
\textbf{Purpose}: The AM-GM inequality specialized for a multiset whose product is at least 1, proving that the sum is bounded below by its cardinality.

\vspace{0.5em}
\textbf{Proof Body Outline}

\begin{enumerate}
  \item Define $P$ as the sum $\sum_{i \in \text{Finset.Icc } 0 k} \text{monomial } i (c_i)$ and $P_R$ as $P.\text{map } (\text{algebraMap } \mathbb{Z} \text{ } \mathbb{R})$.
  \item Apply \texttt{coeff\_of\_sum\_Icc} to assert $P_R.\text{coeff } i = (c_i : \mathbb{R})$ for $i \in \{0, 1, 2, k\}$.
  \item Establish $P_R.\text{roots.card} = k$ using \texttt{card\_roots\_eq\_of\_ncard\_rootSet} and the natural degree injectivity.
  \item Establish that $P_R.\text{splits } (\text{RingHom.id } \mathbb{R})$ follows from \texttt{Polynomial.splits\_iff\_card\_roots}.
  \item Let $s = P_R.\text{roots}$. Invoke Vieta's formulas (\texttt{Polynomial.coeff\_eq\_esymm\_roots\_of\_splits}) to express $c_0$, $c_1$, and $c_2$ in terms of $s.\text{esymm } i$.
  \item Define multisets $Y$ and $Z$ matching the theoretical blueprint. Use \texttt{multiset\_sum\_sq\_eq}, \texttt{multiset\_map\_erase\_prod\_sum}, and \texttt{multiset\_map\_erase\_prod\_esymm\_two} to show that the sum of the squared elements of $Z$ expands algebraically to exactly $(c_1^2 - 2c_0 c_2 : \mathbb{R}) = (K(c) : \mathbb{R})$.
  \item Use \texttt{multiset\_map\_erase\_prod\_prod} to find $Z.\text{prod} = c_k ((-1)^k c_0)^{k-1}$.
  \item Observe that since $c_0$ and $c_k$ are non-zero integers, their algebraic combination $Z.\text{prod}$ represents a non-zero integer, so its square (the product of $W = Z^2$) is $\ge 1$.
  \item Feed $W$ to \texttt{multiset\_sum\_ge\_card\_of\_prod\_ge\_one} to deduce that $W.\text{sum} \ge W.\text{card}$.
  \item Using $W.\text{card} = k$ and $W.\text{sum} = (K(c) : \mathbb{R})$, deduce $(k : \mathbb{R}) \le (K(c) : \mathbb{R})$. Use \texttt{norm\_cast} to translate this back to $(k : \mathbb{Z}) \le K(c)$.
\end{enumerate}

\end{tcolorbox}


\end{document}